\documentclass{article}

\usepackage{microtype}
\usepackage{graphicx}
\usepackage{subfigure}
\usepackage{booktabs} 
\usepackage{float}
\usepackage{stfloats}
\usepackage{capt-of}

\usepackage{hyperref}

\usepackage[accepted]{icml2024_arxiv}
\usepackage{amsmath}
\usepackage{amssymb}
\usepackage{mathtools}
\usepackage{amsthm}
\usepackage{multirow}

\usepackage[capitalize,noabbrev]{cleveref}

\theoremstyle{plain}

\theoremstyle{definition}

\theoremstyle{remark}

\usepackage[textsize=tiny]{todonotes}

\begin{document}

\twocolumn[
\icmltitle{Using Left and Right Brains Together: Towards Vision and Language Planning}



\icmlsetsymbol{equal}{*}
\icmlsetsymbol{corr}{$\dagger$}

\begin{icmlauthorlist}
\icmlauthor{Jun Cen}{sustech,microsoft,hkust,equal}
\icmlauthor{Chenfei Wu}{microsoft,equal}
\icmlauthor{Xiao Liu}{microsoft}
\icmlauthor{Shengming Yin}{microsoft}
\icmlauthor{Yixuan Pei}{xjtu}
\icmlauthor{Jinglong Yang}{sustech,cityuhk}
\icmlauthor{Qifeng Chen}{hkust,corr}
\icmlauthor{Nan Duan}{microsoft,corr}
\icmlauthor{Jianguo Zhang}{sustech,corr}
\end{icmlauthorlist}

\icmlaffiliation{sustech}{Southern University of Science and Technology}
\icmlaffiliation{microsoft}{Microsoft Research Asia}
\icmlaffiliation{hkust}{The Hong Kong University of Science and Technology}
\icmlaffiliation{xjtu}{Xi'an Jiaotong University}

\icmlaffiliation{cityuhk}{City University of Hong Kong}

\icmlcorrespondingauthor{Qifeng Chen}{cqf@ust.hk}
\icmlcorrespondingauthor{Nan Duan}{nanduan@microsoft.com}
\icmlcorrespondingauthor{Jianguo Zhang}{zhangjg@sustech.edu.cn}

\icmlkeywords{Machine Learning, ICML}

\vskip 0.3in
{
\begin{center}
    \centering
    \includegraphics[width=0.99\linewidth]{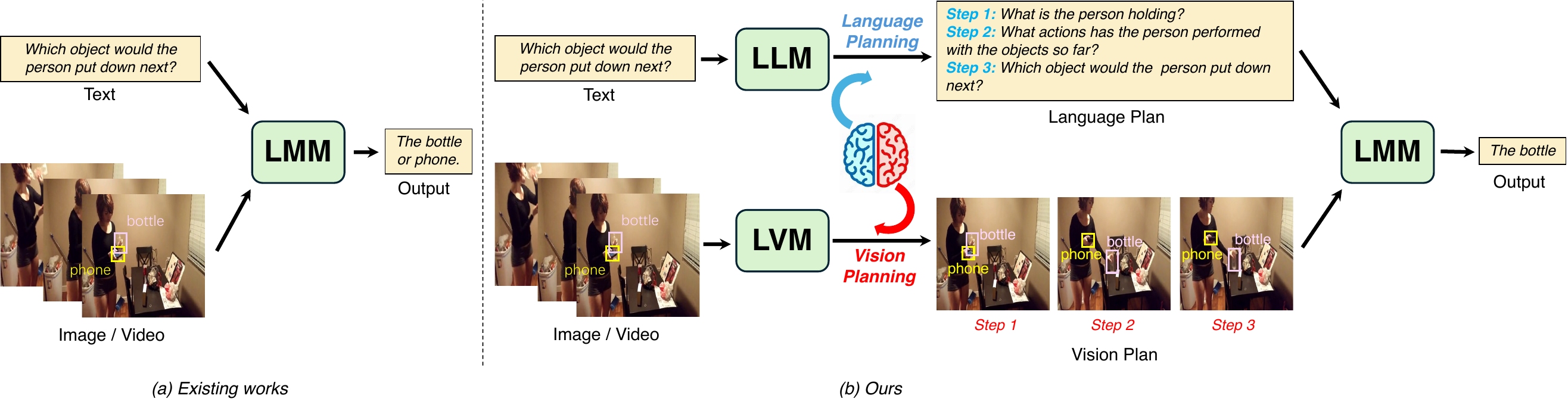}
    \setlength{\abovecaptionskip}{0mm}
    \vspace{-0.5cm}
    \captionof{figure}{Existing Large Multi-modality Models (LMMs) directly make the decision based on text and image inputs. Our Vision-Language Planning (VLP) framework conducts both language planning and vision planning first, which serves as the left hemisphere and the right hemisphere of a human brain, and then use a LMM for the final decision making. }
	\label{fig:tea}
\end{center}
}
]



\printAffiliationsAndNotice{\icmlEqualContribution} 

\begin{abstract}
Large Language Models (LLMs) and Large Multi-modality Models (LMMs) have demonstrated remarkable decision masking capabilities on a variety of tasks. However, they inherently operate planning within the language space, lacking the vision and spatial imagination ability. In contrast, humans utilize both left and right hemispheres of the brain for language and visual planning during the thinking process. Therefore, we introduce a novel vision-language planning framework in this work to perform concurrent visual and language planning for tasks with inputs of any form. Our framework incorporates visual planning to capture intricate environmental details, while language planning enhances the logical coherence of the overall system. We evaluate the effectiveness of our framework across vision-language tasks, vision-only tasks, and language-only tasks. The results demonstrate the superior performance of our approach, indicating that the integration of visual and language planning yields better contextually aware task execution.
\end{abstract}
\section{Introduction}
\label{Intro}

The advent of large-scale auto-regressive text pre-training equips Large Language Models (LLMs) with a powerful ability to conduct sophisticated dialogue and advanced cognitive functions~\cite{brown2020language}. Building upon the strong LLMs, plenty of Large Multi-modality Models (LMMs)~\cite{achiam2023gpt} and agents~\cite{wu2023visual} have been developed to address the multi-modality user demands. These LMMs have shown remarkable achievements across various domains, such as robotics~\cite{du2023learning}, medical diagnosis~\cite{singhal2023large}, and games~\cite{wang2023voyager}.



Most LMMs incorporate a trainable bridge network designed to align visual features with linguistic representations~\cite{liu2023llava}, thereby facilitating the processing of both visual and language tokens by a LLM. Recently, language planning such as Chain-of-Thought (CoT)~\cite{cot, cot_mm} has been integrated into LMMs, offering a structured methodology to decompose intricate questions into more tractable components and enabling a sequenced and step-wise reasoning approach. This kind of CoT language planning has been demonstrated to be effective in both few-shot and zero-shot contexts~\cite{cot, cot_zero}.


Despite the pivotal role of language planning in LMMs, there is a notable shortfall in their capability for vision-based associative reasoning, a process we call vision planning. Language planning alone might result in the generation of responses that are not aligned satisfactorily with the dynamic nature of real-world events, since it is hard to describe the real world with the same granularity and exhaustiveness as visual images by pure language descriptions. In contrast, vision planning could facilitate more realistic reasoning in the form of generating a video that predicts subsequent events using vision inputs. This vision planning is different from the visual branch of current LMMs, which typically maps visual perceptual information into the textual space and still depends on LLMs for linguistic reasoning.

%
From a cognitive perspective, human cognition relies on a symbiotic operation of the brain's hemispheres, with the left primarily governing language and logical reasoning, and the right hemisphere managing spatial awareness and holistic visual intuition~\cite{gazzaniga2005forty,corballis2014left}. For instance, when tackling algebraic mathematical challenges, humans often draw upon geometric interpretations to facilitate the reasoning. Current LLMs exhibit functionalities that are akin to the human left hemisphere, specializing in linguistic processing. Yet, they lack the capacity for visual cognition that is intrinsic to the right hemisphere.


Based on the above observations, we propose a Visual-Language Planning (VLP) framework for multi-modality tasks. With respect to language planning, our approach leverages an LLM such as ChatGPT~\cite{brown2020language} to decompose the input text into several steps which are helpful for responding to the overarching inquiry. With respect to vision planning, we employ a Large Vision Model (LVM) such as Stable Video Diffusion~\cite{blattmann2023stable} to generate future video sequences from current images or videos, maximizing the use of visual information for reasoning that aligns with real-world scenarios. For instance, in Fig.~\ref{fig:tea}, by observing the state of a woman drinking water and holding a cellphone, we generate the subsequent videos where the woman is putting down the bottle. Ultimately, our methodology integrates the outcomes of language and vision planning through an LMM and makes the final decision. Our experiments show the effectiveness of our VLP framework across vision-language tasks, vision-only tasks, and language-only tasks.



\begin{figure*}
  \centering
  \includegraphics[width=0.99\textwidth]{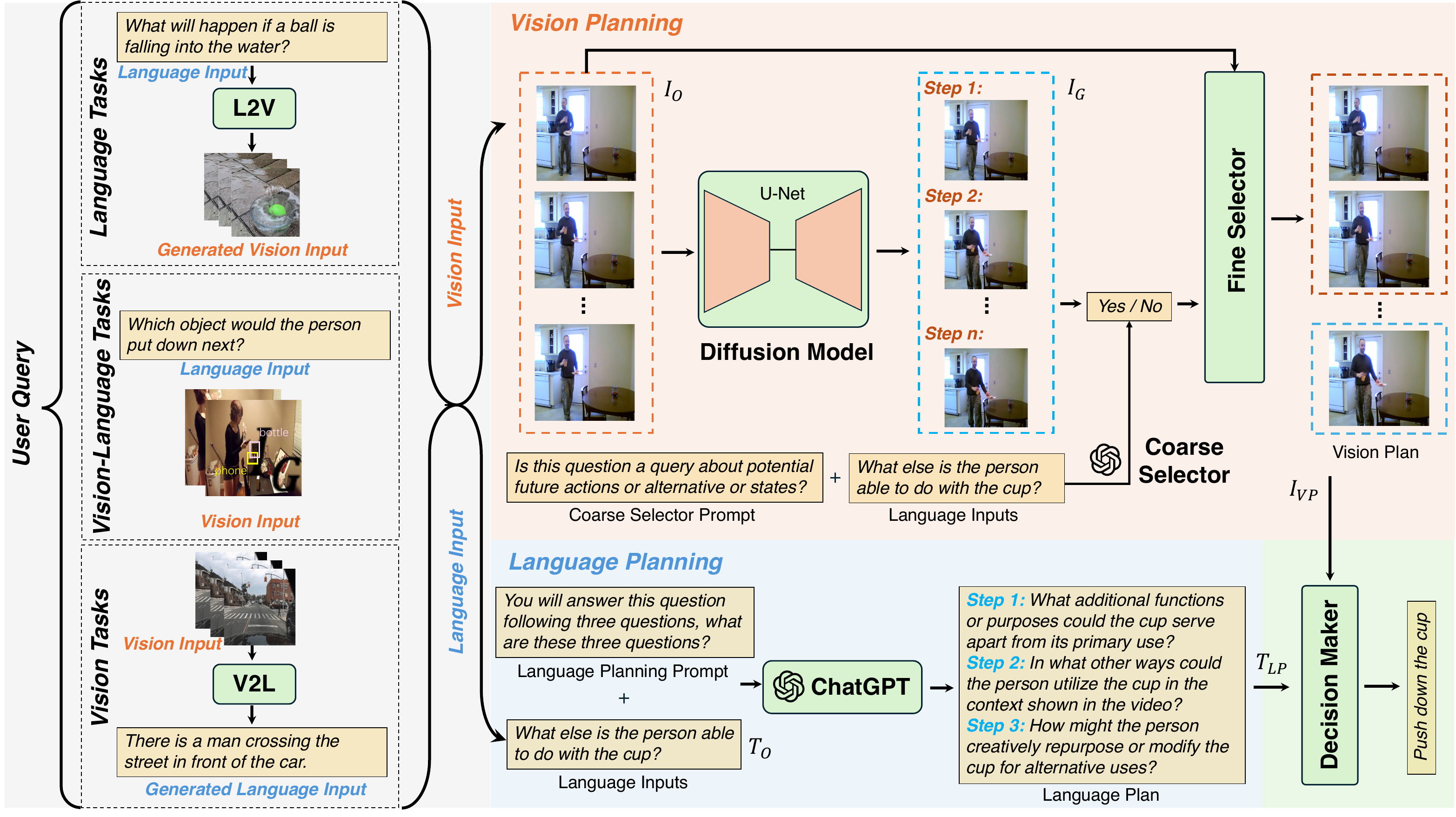}
  \vspace{-0.4cm}
  \caption{Vision-Language Planning (VLP) Framework. We begin by transforming the user queries into the vision input $I_O$ and language input $T_{O}$ for tasks of different modalities. Subsequently, the vision planning and language planning are conducted in parallel to obtain the vision plan $I_{VP}$ and language plan $T_{LP}$. A decision maker then synthesizes these plans to generate the final output.}
  \label{fig:main}
\end{figure*}




In summary, our contributions include the following:
\vspace{-0.3cm}
\begin{itemize}
\setlength\itemsep{0.1em}
\item[$\bullet$] We propose Visual-Language Planning (VLP), a general multi-modal reasoning architecture, which involves not only language planning (serves as left brain) but also vision planning (serves as right brain).

\item[$\bullet$] We implement Visual-Language Processing (VLP) by integrating advanced language generative models, such as ChatGPT, with vision generative models like Stable Video Diffusion, thereby enabling them to collaborate in solving complex problems.

\item[$\bullet$] We show that our VLP not only significantly enhances performance in vision-language tasks but also demonstrates great potential in pure vision and language tasks.
\end{itemize}

\section{Related Work}
\label{related_work}

\subsection{Large Multi-modality Models}
Large Language Models (LLMs) have exhibited impressive capabilities in conversation and reasoning, owing to extensive auto-regressive pre-training methodologies~\cite{brown2020language, touvron2023llama}. Building on the foundation of LLMs, a series of Large Multi-modal Models (LMMs) have been developed, which can process both visual and linguistic inputs~\cite{achiam2023gpt,team2023gemini}. The majority of open-source LMMs employ a strategy that aligns visual features with linguistic representations, and conduct visual instruction tuning to improve performance~\cite{liu2023llava, zhu2023minigpt, team2023internlm}. 
These LMMs make decisions based solely on text and image inputs, which constrains their reasoning abilities. In contrast, our VLP framework initially engages in both language and vision planning, analogous to the left and right hemispheres of the human brain, respectively. An LMM is used finally for the final decision-making process.

\subsection{Planning with Large Language Models}
Most LLMs and LMMs perform planning in the linguistic aspect. The Chain-of-Thought (CoT) approach has been established as an effective technique for prompting LLMs to engage in sequential reasoning~\cite{cot}. Zero-shot CoT~\cite{cot_zero} demonstrates that the prompt ``let's think step by step'' can enhance the model's output without additional effort. In contrast, few-shot CoT~\cite{cot, cot_auto} employs reasoning templates that guide the LLM through to think in a sequential reasoning format. The recent advent of multi-modal CoT~\cite{cot_mm} introduces a two-stage framework that separates rationale generation from answer inference, allowing the latter to fully leverage multi-modal rational information. However, the above works only consider planning in linguistic modality, limiting their capability in visual imagination during planning. Recent studies have employed LMMs in conjunction with video generation models to facilitate task planning in robotics~\cite{du2023learning, du2023video, ajay2023compositional}, where the video generation model functions as a format of visual planning. However, these works only focus on the robotic domain, limiting the exploration in open-domain scenarios. To address this issue, we design a general-purpose VLP that includes both language planning and vision planning and conduct detailed experiments on a variety of downstream tasks, including vision-language tasks, vision-only tasks, and language-only tasks.


\subsection{Video Generation}
Initial video generation methodologies~\cite{tulyakov2018mocogan,skorokhodov2022stylegan,wang2023styleinv} utilized generative adversarial networks (GANs)~\cite{goodfellow2020generative}, yet they were limited in producing high-quality videos~\cite{blattmann2023align}. The advent of diffusion models~\cite{rombach2022high}, characterized by their stable training process and superior generative capabilities, has led to their adoption in contemporary video generation techniques~\cite{ho2022imagen,blattmann2023stable,luo2023videofusion,yin2023nuwa,zhang2023i2vgen}. Among these, Stable Video Diffusion~\cite{blattmann2023stable} has gained recognition for its robust text-to-video and image-to-video generation capabilities across various domains. DMVFN~\cite{hu2023dynamic} tailors video generation to specific applications, such as autonomous driving, by operating on video inputs. Meanwhile, MCVD~\cite{voleti2022mcvd} innovatively masks and reconstructs video frames, facilitating video prediction and interpolation. In our Visual Language Processing (VLP) framework, we integrate a video generation model to augment the visual aspect of the reasoning process.


\section{Vision-Language Planning}
\label{method}



\subsection{Framework Overview}
As shown in Fig.~\ref{fig:main}, Our VLP system handles user queries of different modalities, including pure language tasks, pure vision tasks, and vision-language tasks. For pure language tasks, a Language-to-Vision (L2V) model is used to convert language queries to corresponding visual content, such as images or videos. Conversely, for pure vision tasks, relevant language descriptions are produced using a Vision-to-Language (V2L) model. Therefore, whatever modalities the user queries are, our approach enables the acquisition of both vision input $I_O$ and language input $T_O$.

The vision input $I_O$ undergoes processing by the vision planning branch to yield the vision planning outcomes $I_{VP}$. A video generation diffusion model is employed to synthesize future frames that constitute the vision plan, followed by the use of coarse and fine selectors to choose frames that are potentially beneficial for the current task.
The language input $T_O$ is processed by an LLM to produce the language plan $T_{LP}$.
Finally, a decision maker makes the final decision based on the vision plan $I_{VP}$ and language plan $T_{LP}$.

\subsection{Vision Planning}
\textbf{Visual Planning Generator (VPG).} The vision input is denoted as $I_O=\{I_O^1,I_O^2,...,I_O^N \}$, where $N$ represents the number of input images. $N=1$ means we input an image and $N>1$ means the vision input is a video. Then a Visual Planning Generator (VPG) $G$ is applied to generate the future frames $I_G$:
\begin{equation}
    I_G = G(I_O),
    \label{eq:ge}
\end{equation}
where $I_G=\{I_G^1,I_G^2,...,I_G^n \}$ and $n$ denotes the number of generated images or vision planning steps. The video diffusion model $G$ is an image-to-video model if the input is an image ($N=1$), and $G$ is a video prediction model if the input is a video ($N>1$).

\textbf{Vision Planning Selector (VPS).}
Although VPG generates potentially useful future frames, directly using them may cause the following issues: 1) We notice that not all problems are related to the future states, in which case the inclusion of generated frames could introduce irrelevant noise. 2) Besides, the video generation model's limitations might result in artifacts and superfluous frames within the generated content.
To address above issues, we employ a Vision Planning Selector (VPS) comprising two modules: 1) Coarse Selector (CS) to determine whether the current task needs the generated video frames or not. 2) Fine Selector (FS) to determine which frames should be selected to help solve problems if current task requires generated frames. 

For the Coarse Selector (CS), we simply add the prompt \textit{[Is this question a query about potential
future actions or alternative or states?]} to ChatGPT, so that it will output Yes or No to judge if the language query $T_O$ should use the generated frames or not.
For the Fine Selector (FS), it selects the useful frames for the query $T_O$ among original inputs $I_O$ and generated frames $I_G$ as the ultimate vision plan $I_{VP}$. FS takes a video as the input, and assigns selection scores for each frame, so that we can select the frames with top-$K$ highest scores to form the final vision plan $I_{VP}$.
Specifically, for each frame, FS first extracts visual features by a CLIP vision encoder~\cite{radford2021learning}. Then visual query features are generated by a Q-former~\cite{li2023blip} and concatenated with the text prompt like \textit{[Does the information within the frame provide the necessary details to accurately answer the given question]}~\cite{yu2023self}. Finally, a LLM takes the visual and text tokens as inputs and we use the output probability of the token ``Yes'' as the selection score for the frame.
The final vision plan $I_{VP}$ can be formed as
\begin{equation}
    I_{VP}=\left\{\begin{array}{l}\textit{FS}(\textit{Concate}(I_{O}, I_{G})), \ if \ \textit{CS}(T_O) = \ \textit{Yes,} \\
\textit{FS}(I_{O}), \ if \ \textit{CS}(T_O) = \ \textit{No.} 
\end{array}\right.
\end{equation}


\subsection{Language Planning}
In our language planning branch, we implement the zero-shot chain-of-thought technique to decompose the language input into a series of sub-steps, forming the language plan $T_{LP}$. We use the prompt like \textit{[Imagine that you are trying to answer a Video Q\&A Multi-choice Question. You will firstly watch a video and then answer this question. Question here. You will answer this question following three questions, what are these three questions?]}. ChatGPT will answer 3 steps which could help the decision maker to think step by step and make the decision according to these sub-questions.

For instance, the user asks \textit{[What else is the person able to do with the cup?]} in Fig~\ref{fig:main}. To answer this question, ChatGPT generates the three-step language plan including \textit{[what additional functions could the cup serve, in what other ways could the person utilize the cup, and how might the person repurpose the cup]}. These language plan steps provide complementary information for the initial query and guide the following decision maker to give the final answer from different perspectives.

\subsection{Decision Maker}
\label{sec:dm}

The decision maker is responsible for making the final output according to the vision plan $I_{VP}$ and language plan $T_{LP}$. We design a multi-round conversation strategy to guide the LMM to think sequentially. 1) Vanilla Answering. We directly give the original vision inputs $I_O$ and language inputs $T_L$ to LMM, and prompt LMM to give the vanilla answer. 2) Language Answering. For language plan $T_{LP}$, we first let the LMM answer three language steps one by one, and then give the answer for the original query $T_O$ based on the answers of all steps. 3) Vision Answering. We prompt LMM to give the answer using generated vision plan $I_{VP}$. 4) Voting. We propose a voting mechanism to strengthen the vanilla answering by the language answering and vision answering, since they provide the alternatives from different modality reasoning
perspectives. LMM will evaluate the validity again between the vanilla answer and language answer or vision answer to obtain the voted language answer and vision answer, and finally make the ultimate decision between these two voted answers. See Fig.~\ref{fig:dm} for an example.

\begin{table*}[t]
\caption{Results on video question answering. 
}

\centering
\setlength{\tabcolsep}{2.7mm}
\begin{tabular}{lcccccccccc}
\toprule
\multirow{2}{*}{{Model (\# Frames)}} & \multicolumn{5}{c}{{STAR}} & \multicolumn{4}{c}{{NExT-QA}}\\
\cmidrule(lr){2-6} \cmidrule(lr){7-10}
& {Int.} & {Seq.} & {Pre.} & {Fea.} & {Avg.} & {Tem.} & {Cau.} & {Des.} & {Avg.}\\
\midrule
 {ViperGPT (dense/1fps)} \cite{suris2023vipergpt}  & - & - & - & - & - & - & - & - & {60.0} \\
Flamingo-80B (30) \cite{alayrac2022flamingo}  & - & - & - & - &  39.7 & - & - & - & -\\
VFC (32) \cite{momeni2023verbs}  & - & - & - & - & - & 45.4 & 51.6 & 64.1 & 51.5\\
InternVideo$^*$ (8) \cite{wang2022internvideo} & 43.8 & 43.2 & 42.3 & 37.4 & 41.6 & 43.4 & 48.0 & 65.1 & 49.1 \\
BLIP-2$^{\text{voting}}$ (4) \cite{li2023blip} & 41.8 & 39.7& 40.2 &39.5 & 40.3 & 59.1 & 61.3 & 74.9 & 62.7\\ 
BLIP-2$^{\text{concat}}$ (4) \cite{li2023blip}& 45.5 & 41.8 & 41.8 & 40.0 & 42.2 & 59.7 & 60.8  & 73.8 & 62.4 \\
SEVILA (4) \cite{yu2023self} & 48.3 & 45.0 & 44.4  & 40.8 & 44.6 & \textbf{61.3} & \underline{61.5} & \underline{75.6} & \underline{63.6} \\
LLAVA (4) \cite{liu2023llava}&\underline{49.0}&\underline{47.3}&\underline{45.5}&\underline{47.8}&\underline{47.4} &55.7 &60.6 &74.3 &61.1  \\ \midrule
VLP (4 + 1 (Generated Frame)) &\textbf{52.0} &\textbf{50.1} &\textbf{50.8} &\textbf{49.0} &\textbf{50.5} &\underline{60.5} &\textbf{63.7} &\textbf{76.7} &\textbf{64.7} \\
\bottomrule
\label{tab:main_zs}
\end{tabular}
\vspace{-0.5cm}
\end{table*}

\begin{table}[t]
\centering
\caption{Results of Video Captioning on the BDD-X dataset. 'B', 'C', and 'M' refer to BLEU-4, CIDEr, and METEOR, respectively.}
\setlength{\tabcolsep}{1.5mm}
\label{tab:bddx}
\begin{tabular}{lccc}
\toprule
{Method} & {B}     & {C}    & {M}\\ \midrule
\multicolumn{1}{l}{S2VT \cite{venugopalan2015sequence}}   & 30.2  & 179.8    & 27.5\\
\multicolumn{1}{l}{S2VT++ \cite{venugopalan2015sequence}} & 27.1   & 157.0    & 26.4 \\
\multicolumn{1}{l}{SAA \cite{kim2018textual}} & 31.8    & 214.8    & 29.1\\
\multicolumn{1}{l}{WAA \cite{kim2018textual}} & 32.3    & 215.8    & 29.2\\
\multicolumn{1}{l}{ADAPT \cite{jin2023adapt}}     &   
  34.6 &
  247.5 &
  30.6\\ \midrule
\multicolumn{1}{l}{VLP (Ours)} &\textbf{35.7} &\textbf{256.7} &\textbf{31.1} \\ 
  \bottomrule
\end{tabular}
\vspace{-0.3cm}
\end{table}

\begin{figure*}[t!]
  \centering
\includegraphics[width=0.99\textwidth]{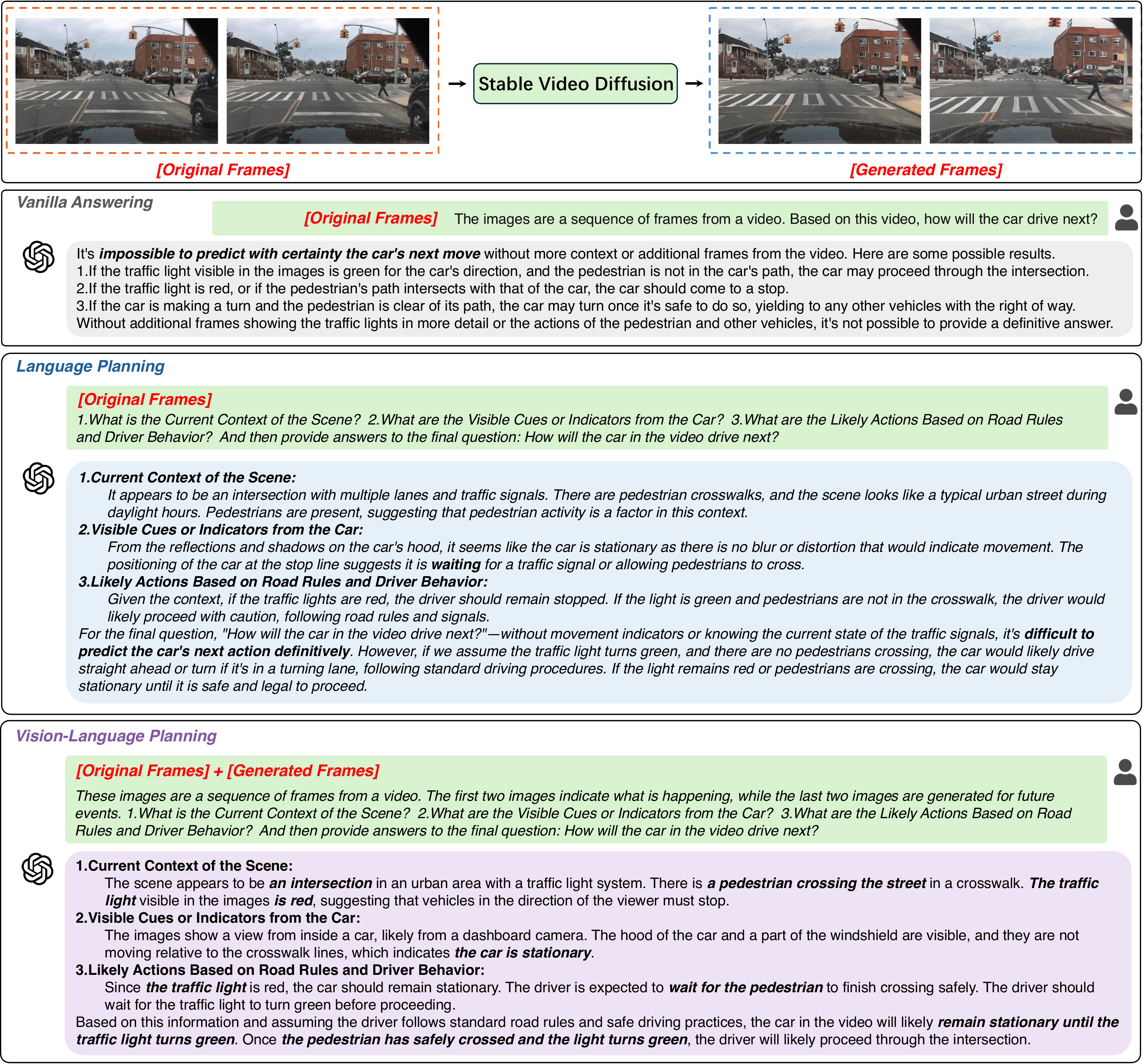}
\vspace{-0.3cm}
  \caption{VLP with GPT4-V on vision-related tasks. Languege planning helps the GPT4-V to consider the information from different perspectives and reason sequentially, and vision planning provides the possible future states which are essential for the ultimate decisions.}
  \label{fig:gpt_4v}
\end{figure*}

\begin{figure*}[t!]
  \centering
\includegraphics[width=0.99\textwidth]{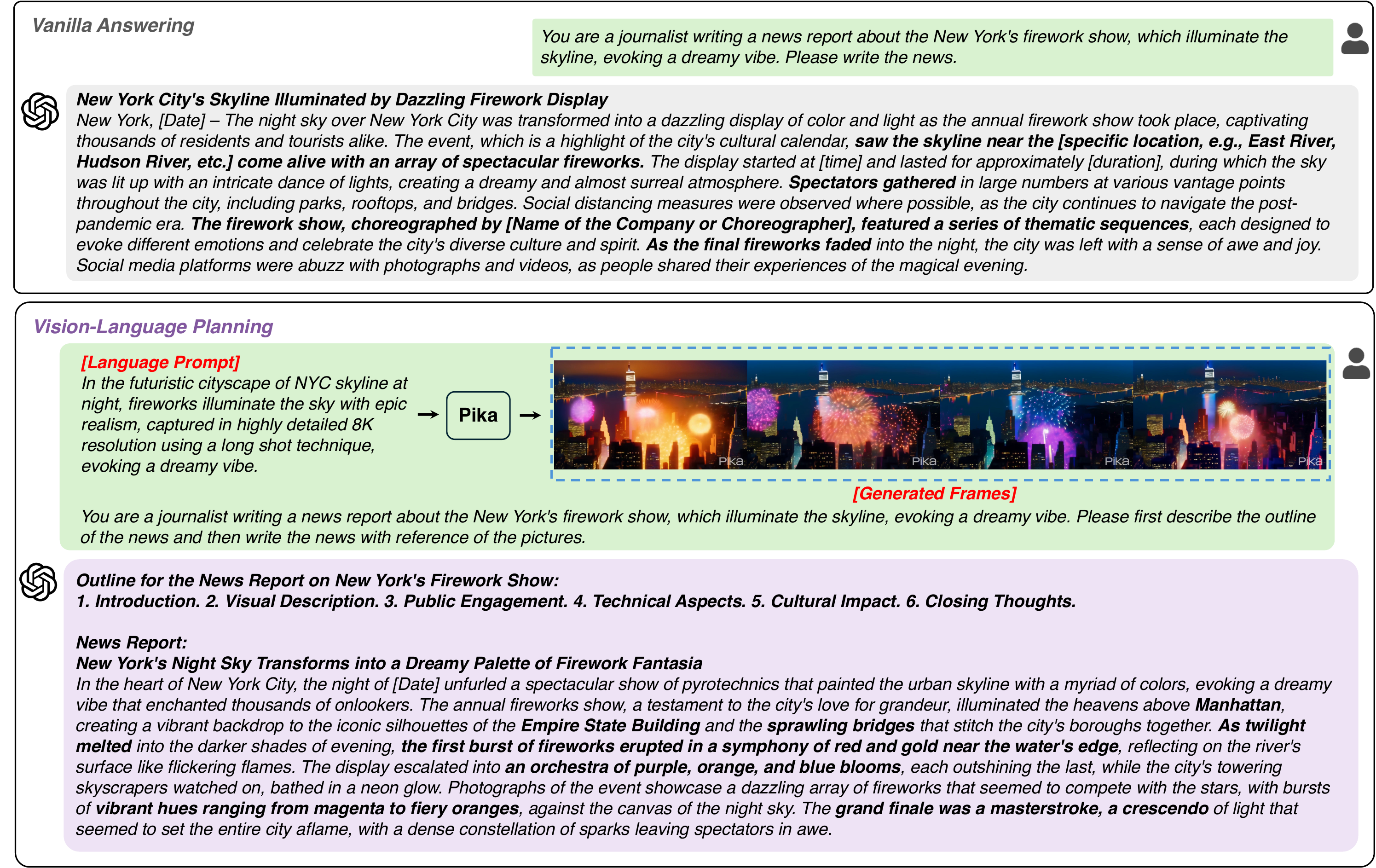}
\vspace{-0.3cm}
  \caption{VLP with GPT4-V on language-only tasks. GPT4-V gives more detailed and vivid descriptions using the generated videos.}
  \label{fig:ex_lp}
\end{figure*}

\section{Experiments}
\subsection{Experimental Settings}
\textbf{Datasets.} We evaluate our VLP on various scenarios, covering the open-domain scenario (STAR~\cite{wu2021star} and NExT-QA~\cite{xiao2021next}), autonomous driving scenario (BDD-X~\cite{kim2018textual}), and robotics operation scenario (BAIR ~\cite{ebert2017self}). 
The dataset details are as following:
\vspace{-0.25cm}
\begin{itemize}
\setlength\itemsep{0.2em}
    \item \textit{STAR.} Situated Reasoning in Real-World Videos (STAR) dataset~\cite{wu2021star} comprises 60k situated reasoning questions accompanied by programs and answers, 24k candidate choices, and 22k trimmed situation video clips. It covers four types of questions: interaction, sequence, prediction, and feasibility, in which prediction and feasibility questions are strongly related to what will happen next. We evaluate the accuracy of the multiple-choice questions.
    \item \textit{NExT-QA. } NExT-QA~\cite{xiao2021next} comprises 5440 videos, each with an average duration of 44 seconds. It includes approximately 52k manually annotated question-answer pairs, categorized into causal (48\%), temporal (29\%), and descriptive (23\%) questions. We evaluate the accuracy of the multiple-choice questions.
    \item \textit{BDD-X.} BDD-X~\cite{kim2018textual} is a textual autonomous driving dataset. It annotates the descriptions and actions of 77 hours within 6,970 videos from BDD dataset~\cite{xu2017end}. The video captioning performance is evaluated by the BLEU-4 score, CIDEr score, and METEOR score. The actions include the course and speed, and we use root mean squared error (RMSE) and a tolerant accuracy ($A_{\sigma}$)~\cite{jin2023adapt} to measure the acition prediction performance.
    \item \textit{BAIR.} BAIR dataset~\cite{ebert2017self} records 30k videos of a robot pushing multiple objects. The X, Y, and Z location of the robot gripper is provided for each frame, and we use root mean squared error (RMSE) for evaluation. We conduct prediction and planning tasks on the BAIR dataset. In the prediction task, we aim to predict the location of the robot gripper of the subsequent frames given the preceding ones. In the planning task, we provide the initial frames and goal (end) frames and predict the locations of the robot gripper between them.
\end{itemize}

\textbf{Implementation Details.} For STAR and NExt-QA dataset, we use Stable Video Diffusion~\cite{blattmann2023stable} model for future frames generation. The Visual Planning Selector (VPS) contains a ChatGPT (Coarse Selector) and a BLIP-2 (Fine Selector)~\cite{li2023blip,yu2023self}. The decision maker is based on a LLAVA model~\cite{liu2023llava}, as shown in Fig.~\ref{fig:dm}. We conduct zero-shot experiments without finetuning LLAVA.

For BDD-X and BAIR datasets, we use DMVFN~\cite{hu2023dynamic} and MCVD~\cite{voleti2022mcvd} for video generation. The Fine Selector of VPS is a temporal transformer following~\cite{wang2023clip}, and the decision maker is a BERT model~\cite{devlin2018bert} based on ADAPT~\cite{jin2023adapt}. We train the decision maker and Fine Selector end-to-end on BDD-X and BAIR datasets, and we follow the training details of ADAPT~\cite{jin2023adapt}.

\subsection{Results}
\textbf{Results on Video Multiple Choice Questions Datasets STAR and NexT-QA.} Video multiple choice questions are typical vision-language tasks, which require the model to have a deep understanding of both the videos and questions. The corresponding results are shown in Table~\ref{tab:main_zs}. It shows that our VLP achieves the best performance among all LMM-based baselines, including the state-of-the-art method SEVILA and our implemented baseline LLAVA. Table~\ref{tab:main_zs} illustrates the effectiveness of our VLP in the open-domain scenario.

\textbf{Results on Video Captioning Datasets BDD-X.} Video captioning is a vision-only input task. Table~\ref{tab:bddx} shows that our VLP surpasses the state-of-the-art method ADAPT with a clear margin.

\textbf{Case Study with GPT4-V.} We cannot conduct quantitative experiments using GPT-4V due to the usage limit restrictions. Instead, we provide two case studies to demonstrate the effectiveness of VLP with GPT4-V. Fig.~\ref{fig:gpt_4v} shows that vanilla answering cannot give the results with current videos (\textit{[It's impossible ...]}). Language planning provides sequential reasoning steps but still \textit{[difficult to predict the next action.]}. With the generated future frames from vision planning, which shows the pedestrian is crossing the road, GPT4-V gives the correct answer that the car should \textit{[remain stationary]} and move \textit{[once the pedestrian has safely crossed and the light turns green]}. Fig.~\ref{fig:ex_lp} shows that our VLP generates more detailed and vivid descriptions based on the generated future frames in language-only tasks. For example, VLP generates the phases like \textit{[fireworks erupted in a symphony of red and gold near the water's edge]} while vanilla answering does not.

\subsection{Ablation Study}
\label{sec:ab}

\textbf{Effects of VP and LP.} We conduct an ablation study of VP and LP on Video Q\&A dataset STAR and Video Captioning dataset BDD-X. Table~\ref{tab:ab_star} and Table~\ref{tab:ab_bddx} show that both VP and LP could clearly boost the performance of the baseline. For example, VP and LP improved 2.2\% and 3.0\% Accuracy on STAR and 1.1 and 0.6 BLEU-4 score on BDD-X. LP brings more benefit than VP on vision-language task STAR while this circumstance is contrary on vision task BDD-X. This is because understanding the language question is also significant for the Q\&A task, while the captioning task has a consistent output demand based on only vision input.

\begin{table}[t]
\caption{Ablation study of VP and LP on STAR dataset. $\dagger$ means using ground truth future frames.
}
\centering
\setlength{\tabcolsep}{1.3mm}
\begin{tabular}{lcccccc}
\toprule
{Model}
& {Int.} & {Seq.} & {Pre.} & {Fea.} & {Avg.}\\
\midrule
LLAVA~\cite{liu2023llava} &49.0&47.3&45.5&47.8 &47.4\\
LLAVA+VP (Ours) &51.5 &\underline{49.9} &50.0 &47.1 &49.6\\
LLAVA+LP (Ours)&\textbf{52.3} &\textbf{50.1} &\textbf{51.1} &\underline{48.2} &\underline{50.4}\\ 
LLAVA+LP+VP (Ours)&\underline{52.0} &\textbf{50.1} &\underline{50.8} &\textbf{49.0} &\textbf{50.5}  \\
\bottomrule
\label{tab:ab_star}
\end{tabular}
\vspace{-0.7cm}
\end{table}

\begin{table}[t]
\setlength{\tabcolsep}{3mm}
\begin{center} 
\caption{Ablation study of VP and LP on BDD-X dataset.}
\label{tab:ab_bddx}
\begin{tabular}{lccc}
\toprule
{Method} & {B}     & {C}    & {M}\\ \midrule
\multicolumn{1}{l}{ADAPT~\cite{jin2023adapt}}     &   
  34.6 &
  247.5 &
  30.6 \\
\multicolumn{1}{l}{ADAPT+LP (Ours)} &35.2 &242.6 &\underline{30.8} \\ 
\multicolumn{1}{l}{ADAPT+VP (Ours)} &\underline{35.7} &\textbf{256.7} &\textbf{31.1} \\ 
\multicolumn{1}{l}{ADAPT+VP+LP (Ours)} &\textbf{36.2} &\underline{251.7} &30.6 \\
  \bottomrule
\end{tabular}
\end{center}
\vspace{-0.2cm}
\end{table}

\begin{table}[t]
\caption{Ablation study of VPS (including CS and FS) on STAR.
}
\centering
\setlength{\tabcolsep}{2.5mm}
\begin{tabular}{lcccccc}
\toprule
{Model}
& {Int.} & {Seq.} & {Pre.} & {Fea.} & {Avg.}\\
\midrule
VP w/o CS &49.6&47.7&\underline{48.6}&\underline{45.9} &\underline{48.0}\\
VP w/o FS &\underline{51.4} &\textbf{50.3} &38.1 &42.4 &45.6\\
VP &\textbf{51.5} &\underline{49.9} &\textbf{50.0} &\textbf{47.1} &\textbf{49.6}\\
\bottomrule
\label{tab:ab_vpd}
\end{tabular}
\vspace{-0.7cm}
\end{table}

\begin{table}[t]
\caption{Ablation study of voting in decision maker on STAR.
}
\centering
\setlength{\tabcolsep}{2.5mm}
\begin{tabular}{lcccccc}
\toprule
{Model}
& {Int.} & {Seq.} & {Pre.} & {Fea.} & {Avg.}\\
\midrule
VP w/o voting &51.4 &\textbf{50.4} &48.1 &43.3 &48.3\\
VP &\textbf{51.5} &49.9 &\textbf{50.0} &\textbf{47.1} &\textbf{49.6} \\ \midrule
LP w/o voting &48.3 &49.8 &44.2 &42.9 &46.3\\
LP &\textbf{52.3} &\textbf{50.1} &\textbf{51.1} &\textbf{48.2} &\textbf{50.4}\\
\bottomrule
\label{tab:ab_vo}
\end{tabular}
\vspace{-0.5cm}
\end{table}


\textbf{Effects of VPS in VP.} Coarse Selector (CS) is to determine whether the generated video is needed for the current task. The \textit{Interaction} and \textit{Sequence} questions in STAR are not supposed to be related to the future frames, and Table~\ref{tab:ab_vpd} shows the performance of them drops about 2\% without CS, which means introducing generated frames might bring noisy information for questions independent of the future. Most of the \textit{Prediction} and \textit{Feasibility} questions are related to the future states so they will be chosen by CS to use generated future frames. Without FS, the performance of \textit{Prediction} and \textit{Feasibility} questions drop dramatically, which illustrates the significance of using FS for picking up useful and high-quality generated frames.

\textbf{Effects of Voting in Decision Maker.} The generated language plan and vision plan may not always be reliable due to the limited ability of the language and video generation model. Table~\ref{tab:ab_vo} shows that letting the model vote again between the vanilla answer and the answer with language or vision plan could effectively enhance the performance.

\textbf{VP for Action Prediction and Planning.} In addition to the language output tasks including video Q\&A and captioning, we also implement VP on the action model. Table~\ref{tab:sensor accuracy} shows that with the help of generated future frames, the model could predict the course and speed more accurately in the driving scenario. We also conduct robotics gripper trajectory prediction (predict the future actions given initial states) and planning (generate the future actions given initial and goal states). Table~\ref{tab:ab_bair_pred} shows that VP also helps in this application.

\begin{table*}[t]
\small
\setlength{\tabcolsep}{1mm}
\begin{center}
\caption{Control Signals Prediction Accuracy on BDD-X dataset.}
\label{tab:sensor accuracy}
\begin{tabular}{@{}ccccccccccccc@{}}
\toprule
\multirow{2}{*}{{Method}} & \multicolumn{6}{c}{{Course}}                               & \multicolumn{6}{c}{{Speed}}                    \\ \cmidrule(l){2-13} 
                        & RMSE(degree)$\downarrow$ & $A_{0.1}\uparrow$ & $A_{0.5}\uparrow$ & $A_{1.0}\uparrow$         & $A_{5.0}\uparrow$ & $A_{10.0}\uparrow$ & RMSE(m/s)$\downarrow$ & $A_{0.1}\uparrow$ & $A_{0.5}\uparrow$ & $A_{1.0}\uparrow$         & $A_{5.0}\uparrow$ & $A_{10.0}\uparrow$ \\ \midrule
Single              & 6.3 & 8.3   & 84.7  & \textbf{90.5} & 97.2  & 98.7   & 3.4       & 5.0   & 25.5  & 37.8  & 86.8  & 98.7   \\
ADAPT &
  6.4 &
  62.2 &
  85.5 &
  89.9 &
  97.2 &
  \textbf{98.8} &
  2.5 &
  11.1 &
  28.1 &
  45.3 &
  94.3 &
  99.5 \\
  ADAPT + VP &\textbf{6.2} &\textbf{65.5} &\textbf{86.2} &90.3 &\textbf{97.3} &\textbf{98.8} &\textbf{2.3} &\textbf{16.1} &\textbf{35.3} &\textbf{51.8} &\textbf{95.2} &\textbf{99.6} \\
  \bottomrule
\end{tabular}
\end{center} 
\vspace{-0.6cm}
\end{table*}

\begin{table}[t]
\setlength{\tabcolsep}{1.5mm}
\begin{center}
\caption{Action Prediction (2+0, 4+0) and Planning (1+1, 1+2) RMSE(cm) on BAIR. i and e refer to initial and end (goal) frames.}
\label{tab:ab_bair_pred}
\begin{tabular}{@{}ccccccccccccc@{}}
\toprule
\# Inputs &Method &X &Y &Z &Sum\\ \midrule
\multirow{2}{*}{2 (i) + 0 (e)} &Baseline &8.75 &7.24 &3.86 &19.85
\\
&Baseline + VP &\textbf{8.68} &\textbf{6.83} &\textbf{3.84} &\textbf{19.36}
\\ \midrule
\multirow{2}{*}{4 (i) + 0 (e)} &Baseline &8.06
&6.70
&\textbf{3.63}
&18.39
\\
&Baseline + VP 
&\textbf{7.72}
&\textbf{6.47}
&3.68
&\textbf{17.86}
\\ \midrule
\multirow{2}{*}{1 (i) + 1 (e)} &Baseline 
&5.74
&5.67
&3.42
&14.83
\\
&Baseline + VP
&\textbf{5.48}
&\textbf{5.46}
&\textbf{3.40}
&\textbf{14.34}
\\ \midrule
\multirow{2}{*}{1 (i) + 2 (e)} &Baseline
&5.54
&\textbf{5.45}
&3.41
&14.39
\\
&Baseline + VP
&\textbf{5.05}
&5.46
&\textbf{3.35}
&\textbf{13.85}
\\
 \bottomrule
\end{tabular}
\end{center} 
\vspace{-0.5cm}
\end{table}

\begin{table}[t]
\caption{Results of different numbers of generated frames on STAR. $\dagger$ means using ground truth future frames.
}
\centering
\setlength{\tabcolsep}{1.5mm}
\begin{tabular}{ccccccc}
\toprule
{\# Generated Frames}
& {Int.} & {Seq.} & {Pre.} & {Fea.} & {Avg.}\\
\midrule
1 &\textbf{51.5} &\textbf{49.9} &\textbf{50.0} &47.1 &49.6\\
2 &\textbf{51.5} &\textbf{49.9} &49.8 &\textbf{47.6} &\textbf{49.7}\\ 
3 &51.4 &\textbf{49.9} &\textbf{50.0} &47.3 &\textbf{49.7}\\ \midrule
$\text{1}^\dagger$&\textbf{51.3} &\textbf{50.6} &\textbf{57.5} &51.4 &52.7 \\
$\text{2}^\dagger$
&\textbf{51.3}
&50.5
&55.5
&\textbf{54.9}
&\textbf{53.0}
\\
$\text{3}^\dagger$
&51.2
&50.5
&48.2
&50.4
&50.1
\\
\bottomrule
\label{tab:ab_star_ge}
\end{tabular}
\vspace{-0.7cm}
\end{table}

\begin{table}[t]
\setlength{\tabcolsep}{3.5mm}
\begin{center} 
\caption{Ablation Study of video generation model on BDD-X.}
\label{tab:ab_ge}
\begin{tabular}{@{}ccccccc@{}}
\toprule
{Video Generation Method} & {B}     & {C}    & {M}\\ \midrule
\multicolumn{1}{l}{MCVD - Cityscapes}   &31.2   &195.3     &26.8 \\
\multicolumn{1}{l}{DMVFN - Cityscapes} &35.0&230.1&\textbf{29.4} \\
\multicolumn{1}{l}{DMVFN - Kitti} & \textbf{35.2} &\textbf{234.2} &\textbf{29.4}\\
\multicolumn{1}{l}{Stable Video Diffusion} &33.9 &229.6 &28.8 \\ \midrule 
Ground Truth Frames &34.6 &247.5 &30.6
\\
  \bottomrule
\end{tabular}
\end{center}
\vspace{-0.7cm}
\end{table}

\begin{table}[t]
\setlength{\tabcolsep}{0.5mm}
\begin{center} 
\caption{Results of different numbers of generated frames on BDD-X.}
\label{tab:ab_bddx_ge}
\begin{tabular}{lcccccc}
\toprule
{\# Generated Frames} & 2     & 4    & 8 &16 &30\\ \midrule
\multicolumn{1}{c}{BLEU-4}   &   
  32.0 &
  33.5 &
  \textbf{35.2} &34.4 &33.3 \\
\multicolumn{1}{c}{CIDEr} &212.6 &216.8 &\textbf{234.2} &228.0 &223.3\\ 
\multicolumn{1}{c}{METEOR}&{29.0} &{29.3} &\textbf{29.4} &29.2 &28.6\\ 
  \bottomrule
\end{tabular}
\end{center}
\vspace{-0.5cm}
\end{table}

\textbf{Video Generation Quality Matters.}
The video generation quality plays a significant role in our visual planning. Table~\ref{tab:ab_star_ge} shows that using real future frames has significantly better performance than generated frames using Stable Video Diffusion, \textit{e.g.}, 57.5 and 54.9 compared to 50.0 and 47.6 on \textit{Prediction} and \textit{Feasibility} questions. Fig.~\ref{fig:ex_star_vp_ab} and Fig.~\ref{fig:ex_star_vp_gt} give cases where the generated contents are not reasonable enough to provide positive information while ground truth future frames are helpful. Due to the limited quality of the generated future frames in the open domain, selecting more frames does not have clear performance improvements according to Table~\ref{tab:ab_star_ge}.

On the BDD-X dataset, we select the first 2 frames out of all 30 frames as the input to conduct the ablation study. Table~\ref{tab:ab_ge} shows using ground truth future frames achieves better overall performance than generated frames. MCVD performs worst since it generates low-resolution images. DMVFN trained on the driving datasets including Cityscapes~\cite{cordts2016cityscapes} and Kitti~\cite{geiger2013vision} show better performance because of higher resolution. Stable Video Diffusion does not perform better as it is not specifically trained for the driving scenario. Table~\ref{tab:ab_bddx_ge} shows that a proper number of generated frames is helpful when using domain-specific generative models, but long sequence generated videos are not reliable enough.
\section{Conclusion}

In conclusion, we propose a Visual-Language Planning (VLP) framework in this work. By incorporating both vision-based associative reasoning and language planning, our VLP framework has demonstrated enhanced capabilities in handling multi-modality tasks, which aligns with the cognitive processing strategies of humans involving both hemispheres of the brain. We hope our work could inspire the community to develop more advanced and human-like artificial intelligence systems.

\bibliography{example_paper}
\bibliographystyle{icml2024}

\newpage
\appendix
\onecolumn
\section{Decision Maker}
\begin{figure*}[h!]
  \centering
  \includegraphics[width=0.99\textwidth]{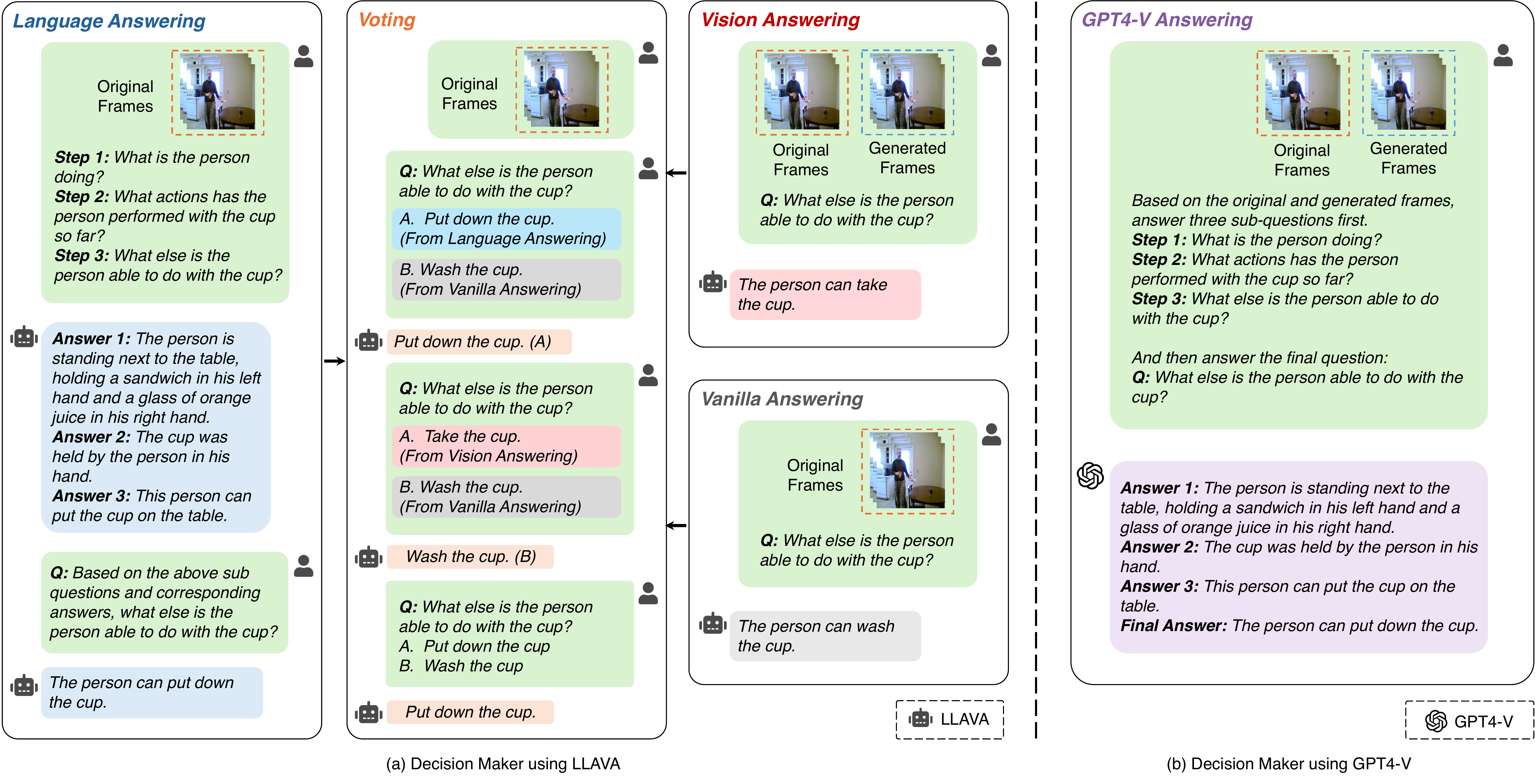}
  \caption{Decision maker using open-sourced LMM like LLAVA and GPT4-V.}
  \label{fig:dm}
\end{figure*}

We give a detailed example of our proposed LMM-based decision maker in Fig.~\ref{fig:dm}. For LMM that has very strong visual instruction following ability like GPT4-V, it could directly answer the overall question following the language plan and vision plan, as shown in Fig.~\ref{fig:dm} (b). However, we find that the open-source LMM such as LLAVA can only follow simple visual instructions and cannot handle flexible and complicated visual instructions. For example, LLAVA cannot answer the questions sequentially in one reply. So we design a multi-round conversation strategy, as shown in Fig.~\ref{fig:dm} (a), which is explained in Sec.~\ref{sec:dm}. Either vision planning and language planning could introduce the noise and our voting mechanism could effectively strengthen the robustness of the final answer, which is testified by Table~\ref{tab:ab_vo}.

\section{VLP Case Study with GPT4-V}
\begin{figure*}[!h]
  \centering
  \includegraphics[width=0.99\textwidth]{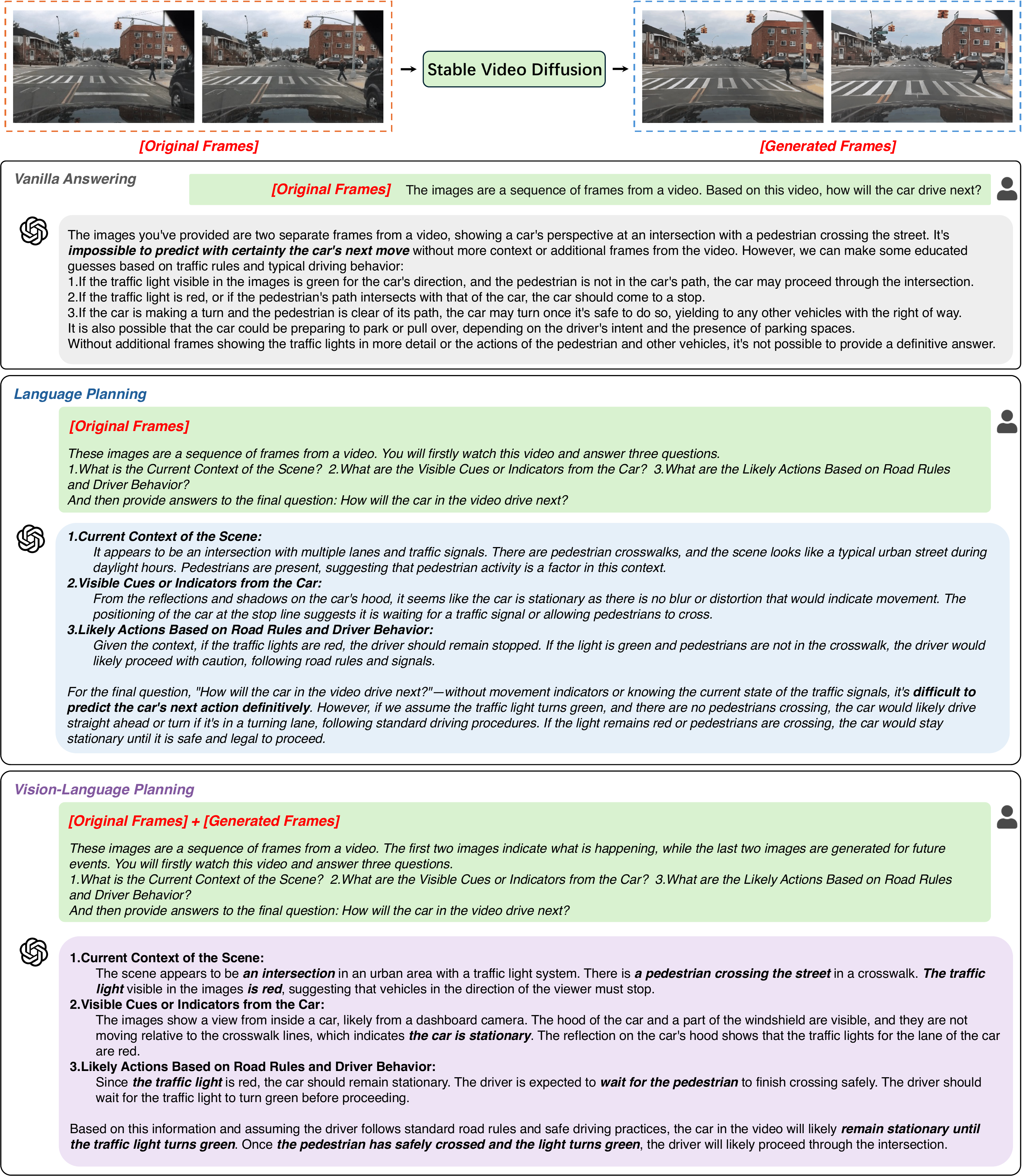}
  \caption{VLP with GPT4-V on vision-related tasks.}
  \label{fig:case_gpt4v_v1}
\end{figure*}

\begin{figure*}[!h]
  \centering
  \includegraphics[width=0.99\textwidth]{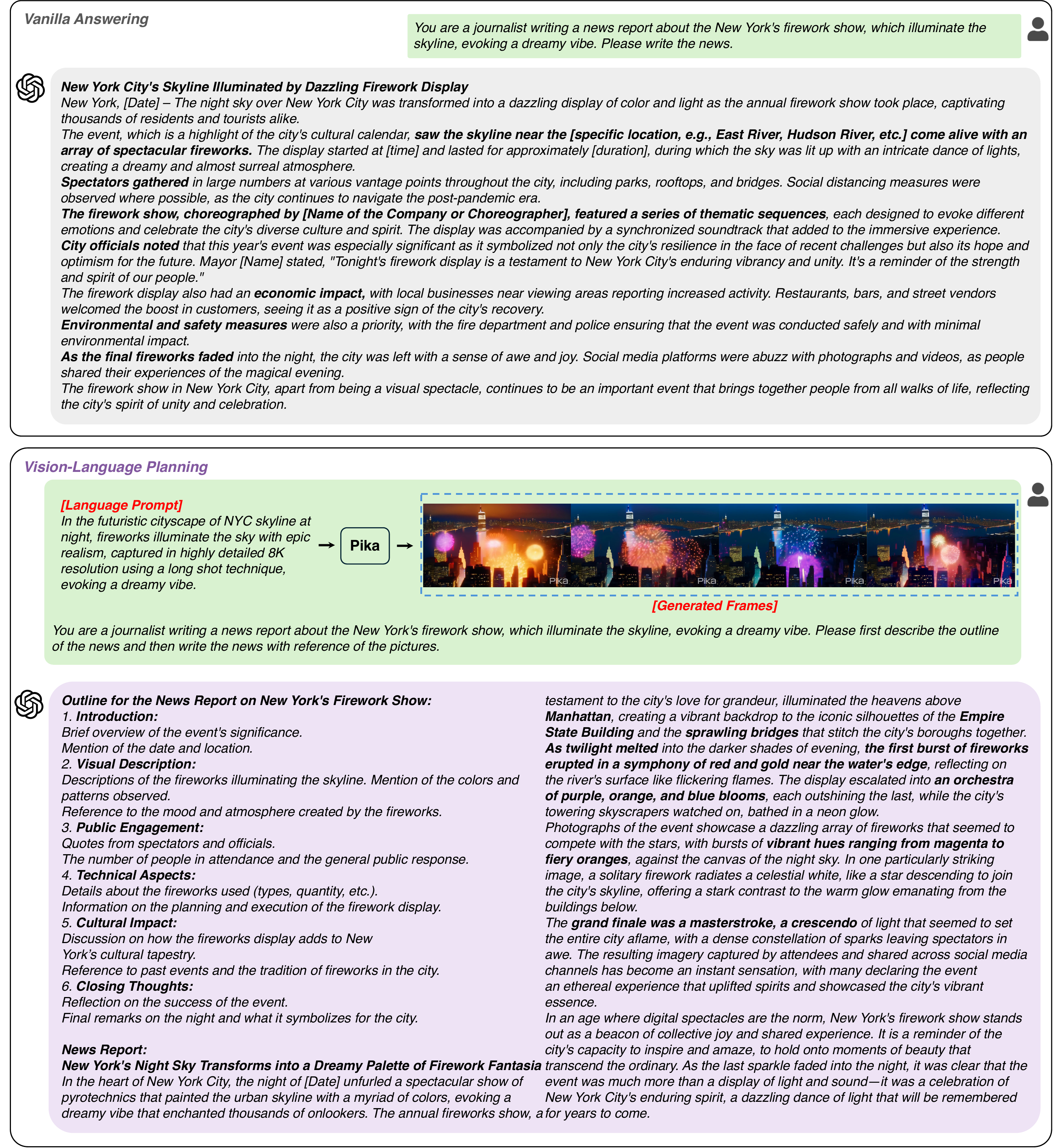}
  \caption{VLP with GPT4-V on language-only tasks.}
  \label{fig:case_gpt4v_l1}
\end{figure*}

We provide the case study of vision-related task in Fig.~\ref{fig:case_gpt4v_v1} and language-only task in Fig.~\ref{fig:case_gpt4v_l1}. 

Fig.~\ref{fig:case_gpt4v_v1} is the detailed version of Fig.~\ref{fig:gpt_4v}. The vanilla answering gives a general and ambiguous answer and requires more information for the decision making. It is \textit{[impossible to predict]} the next move based on the current condition. Language planning decomposes the question into three sub-questions, and let the model answer these sub-questions one by one. Although more information is obtained through language planning, it is still \textit{[difficult to predict]} the next move. Then we use the Stable Video Diffusion for vision planning to generate the future frames, which show that the pedestrian is crossing the road. Using this vision plan, GPT4-V gives the ideal answer that the car should proceed \textit{[once the pedestrian has safely crossed and the light turns green]}.

Fig.~\ref{fig:case_gpt4v_l1} is the detailed version of Fig.~\ref{fig:ex_lp}. The language planning provides a specific outline for the answer, resulting in a better-structured long article. For example, GPT4-V shows that the result should follow \textit{[Introduction, Visual Description, Public Engagement, etc.]}. Besides, it shows that the answer of GPT4-V is more vivid and realistic because of the introducing of generated videos. For example, GPT4-V gives specific location such as \textit{[Empire
State Building and the sprawling bridges]} and detailed description of fireworks like \textit{[the first burst of fireworks
erupted in a symphony of red and gold near the water's edge]} and \textit{[an orchestra
of purple, orange, and blue blooms]}.

\clearpage
\section{VLP Case Study with LLAVA}

We provide several case studies with LLAVA from STAR dataset. Fig.~\ref{fig:ex_star_vp_ab} and Fig.~\ref{fig:ex_star_lp_ab} show the successful and unsuccessful cases of vision planning and language planning. Fig.~\ref{fig:ex_star_vp_gt} shows the vision planning using ground truth future frames.

\subsection{Vision Planning Case Study}

\begin{figure*}[!h]
  \centering
\includegraphics[width=0.99\textwidth]{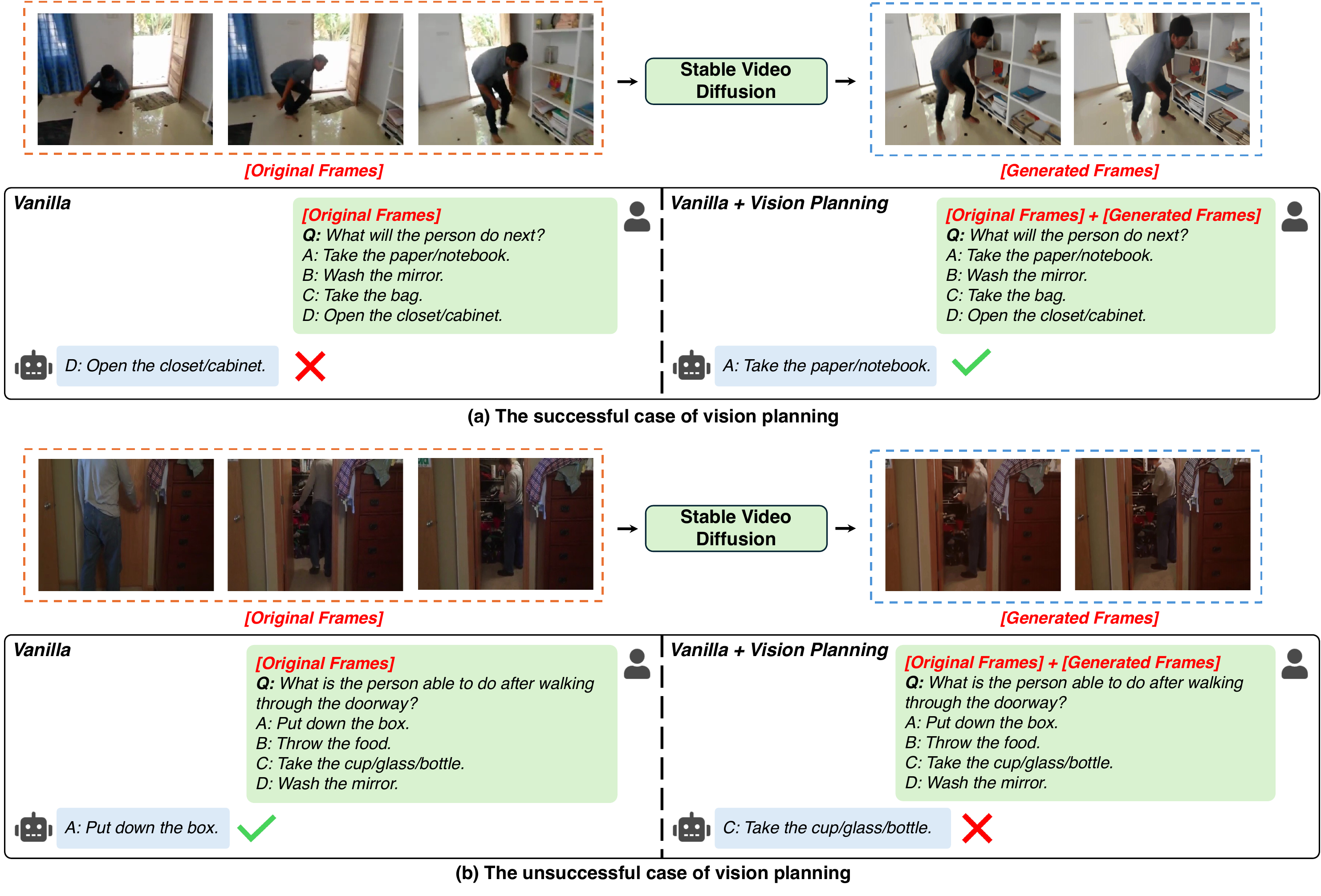}
  \caption{The successful and unsuccessful cases of vision planning.}
  \label{fig:ex_star_vp_ab}
\end{figure*}

Fig.~\ref{fig:ex_star_vp_ab} (a) shows that LLAVA thinks the man is going to open the cabinet without vision planning, which is reasonable according to the background and the action of the man in the video. Stable Video Diffusion generates the future frames which show that the man is reaching out his hand to the paper, so LLAVA gives the correct answer. In Fig.~\ref{fig:ex_star_vp_ab} (b), LLAVA gives the correct answer with original frames since there is a white box in the man's hand (please zoom in the figure for better visualization). However, the generated future frames show the man continue turning around and does not put down the box, so LLAVA gives the wrong answer.

\clearpage
\subsection{Language Planning Case Study}
\begin{figure*}[!h]
  \centering
\includegraphics[width=0.99\textwidth]{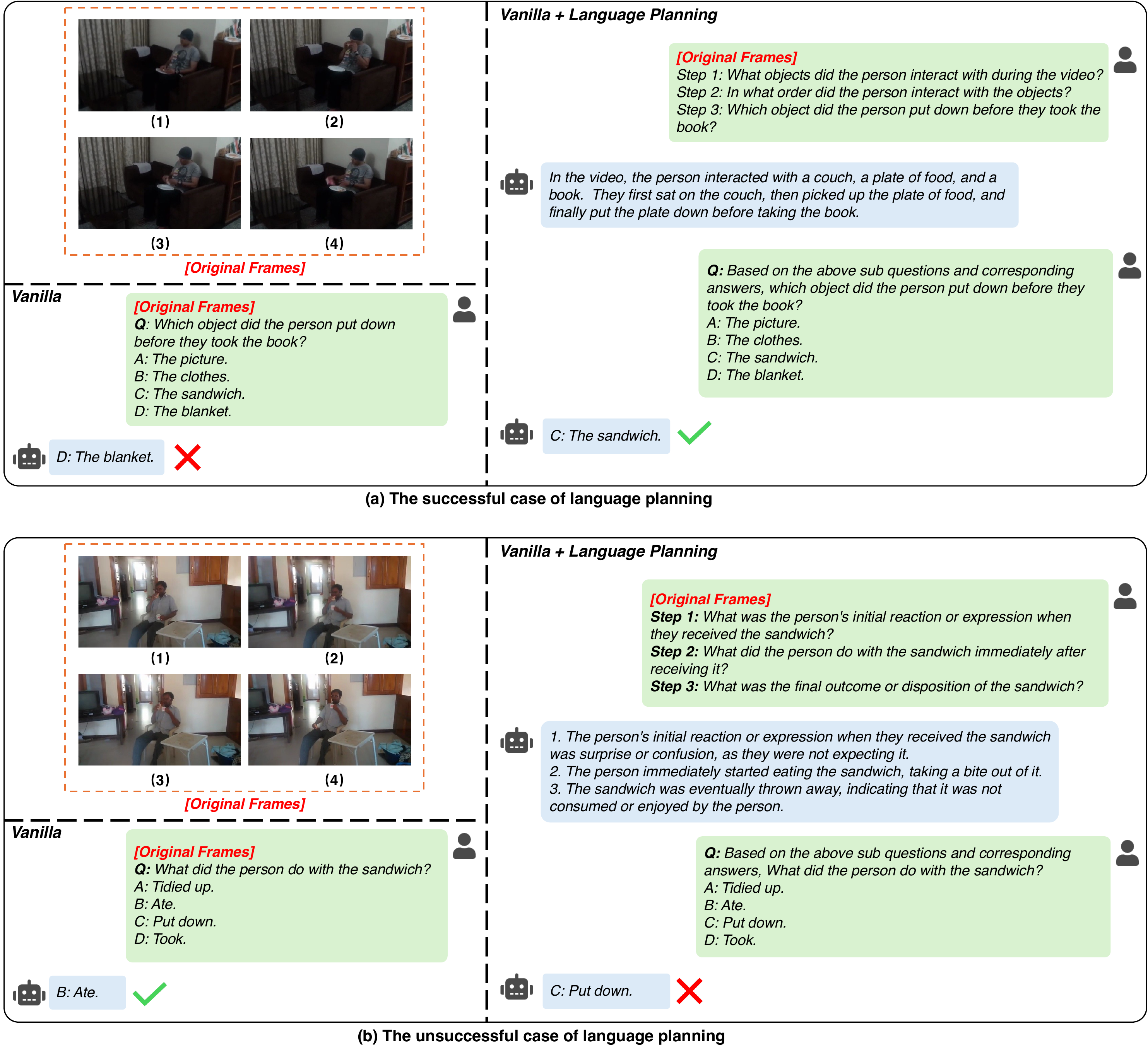}
  \caption{The successful and unsuccessful cases of vision planning.}
  \label{fig:ex_star_lp_ab}
\end{figure*}

Fig.~\ref{fig:ex_star_lp_ab} (a) shows that LLAVA believes that the person puts down the blanket before he took the book, and we believe that the reason is the blanket is prominent in the images. Using Language planning, LLAVA first answers three sub-questions and mentions the man \textit{[puts the plate down before taking the book]}. The sandwich is the only choice that is related to the plate, so LLAVA chooses sandwich and gives the correct answer.

In Fig.~\ref{fig:ex_star_lp_ab} (b), LLAVA gives the correct answer for the question without language planning. However, LLAVA gives the wrong responses when answering the sub-questions of language planning, \textit{e.g.}, the person is \textit{[not expecting it]} and \textit{[thrown away]} the sandwich. Therefore, the final answer picked up by LLAVA is putting down the sandwich which is directly related to \textit{[thrown away]}. This case shows that the quality of generated language plans and corresponding answers for the language plans are significant.

\clearpage
\subsection{Vision Planning using Ground Truth Future Frames}

\begin{figure*}[!h]
  \centering
\includegraphics[width=0.99\textwidth]{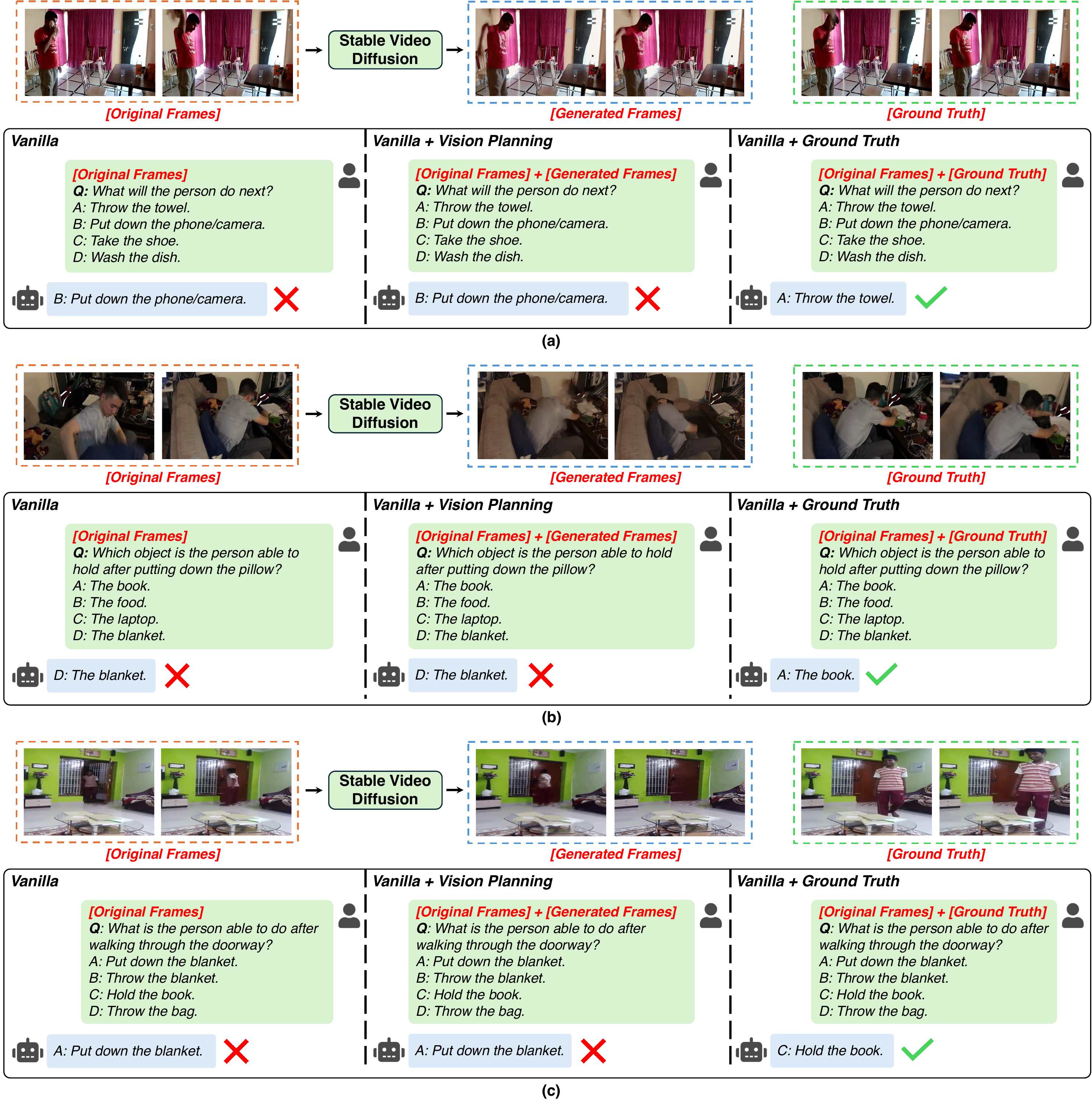}
  \caption{The successful cases of vision planning using ground truth future frames.}
  \label{fig:ex_star_vp_gt}
\end{figure*}

Current LLMs and LMMs have strong text generation ability across open domains, but the video generation models are still far behind in terms of generation ability. In Fig.~\ref{fig:ex_star_vp_gt} (a), the generated arm action is not reasonable enough. In Fig.~\ref{fig:ex_star_vp_gt} (b), the moving part of the generated video is blurry. In Fig.~\ref{fig:ex_star_vp_gt} (c), the person disappears without opening the door. For these cases, using generated future frames cannot correct mistakes for LLAVA but using ground truth future frames can. These cases show that the ability of video generation model is a bottleneck for vision planning, which has been discussed in the last part of Sec.~\ref{sec:ab}.

\clearpage
\section{VLP Case Study with BERT}

Our VLP framework can be utilized not only with recent LLMs and LMMs, but it can also be applied to the traditional BERT for captioning task (Fig.~\ref{fig:ex_bddx} from BDD-X dataset) and action generation task (Fig.~\ref{fig:ex_bair_pre} and Fig.~\ref{fig:ex_bair_int} from BAIR dataset).

\subsection{Video Captioning Case Study}

\begin{figure*}[!h]
  \centering
  \includegraphics[width=0.99\textwidth]{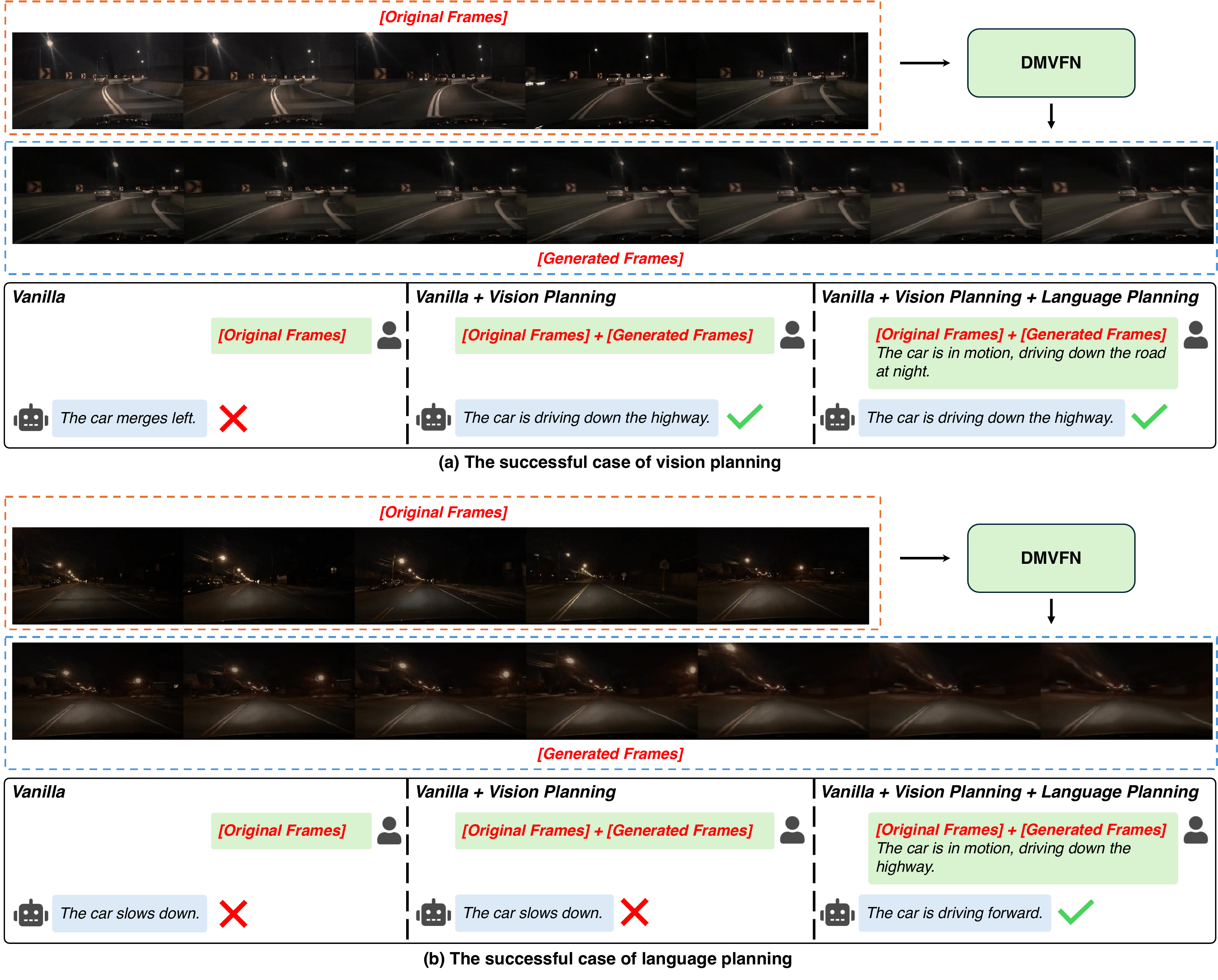}
  \vspace{-0.5cm}
  \caption{Vision Planning and language planning for the video captioning task.}
  \label{fig:ex_bddx}
\end{figure*}

Fig.~\ref{fig:ex_bddx} (a) shows that BERT model predicts the car merges left which is contradictory to the truth that the car is merging right and driving down the highway. With generated future frames, BERT model gives the correct answer. In Fig.~\ref{fig:ex_bddx} (b), both vanilla and vision planning do not give the correct answer, while language planning provides the optimal response with the hint from generated language descriptions.

\subsection{Action Generation Case Study}

\begin{figure*}[!h]
  \centering
\includegraphics[width=0.9\textwidth]{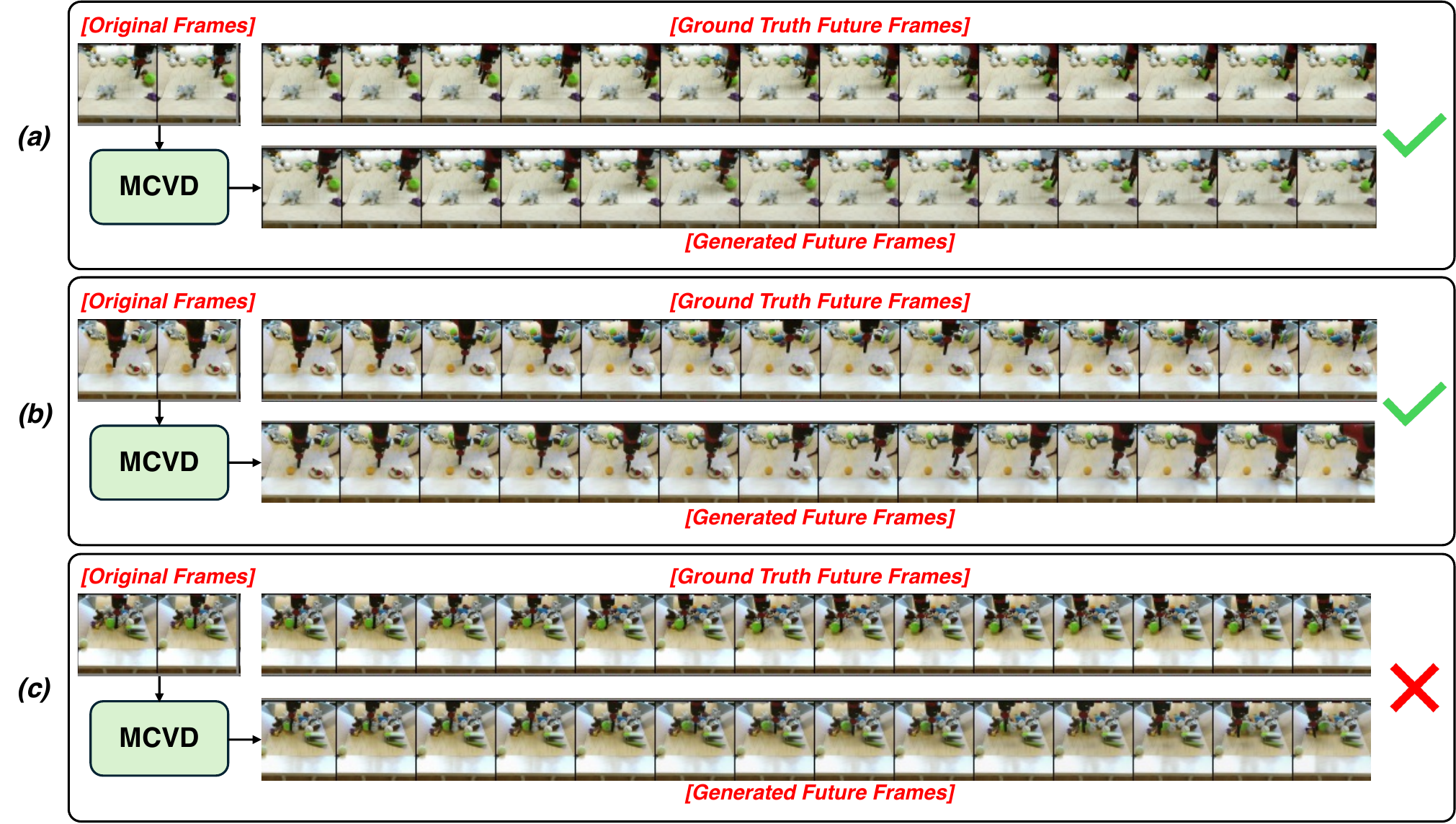}
\vspace{-0.5cm}
  \caption{Action prediction with vision planning.}
  \label{fig:ex_bair_pre}
  \vspace{-0.5cm}
\end{figure*}

\begin{figure*}[!h]
  \centering
\includegraphics[width=0.9\textwidth]{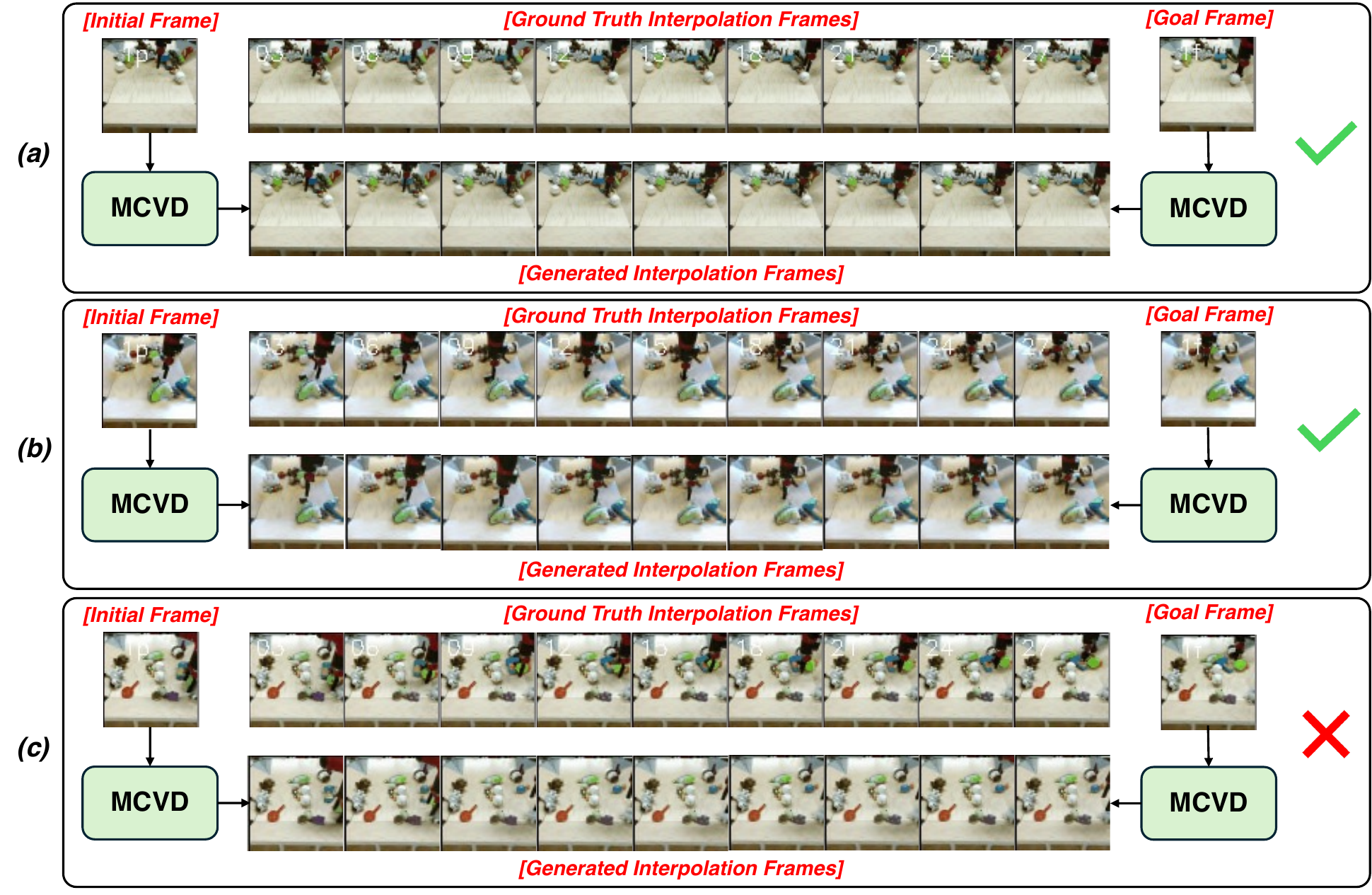}
\vspace{-0.5cm}
  \caption{Action planning with vision planning.}
  \label{fig:ex_bair_int}
\end{figure*}

We provide several cases for action prediction (predict the next actions based on first two frames) and action planning (predict the actions between the initial frame and end frame) in Fig.~\ref{fig:ex_bair_pre} and Fig.~\ref{fig:ex_bair_int}, respectively. 

Fig.~\ref{fig:ex_bair_pre} (a) and (b) show that the generated vision plans successfully predict the gripper to grab the green ball and leave the yellow ball, but Fig.~\ref{fig:ex_bair_pre} (c) shows that the gripper in the generated future frames circles around the green ball without makes contact, which is not the original intention in the ground truth. 

Fig.~\ref{fig:ex_bair_int} shows that the inference results of the planning task are closer to the ground truth than that of the prediction task, since the goal state is given and the video generation process is guided by the goal. Fig.~\ref{fig:ex_bair_int} (a) shows that the generated video successfully predicts the gripper
to approach the white ball. Fig.~\ref{fig:ex_bair_int} (b) is a more sophisticated task, where the gripper first approaches the black object, and then moves the black object up, and finally leaves the black object. The generated video successfully reproduces the whole process according to the final position of the black object, which shows the great potential of vision planning. In Fig.~\ref{fig:ex_bair_int} (c), the video generation model fails to generate the correct process that the gripper is moving the green ball. 

\end{document}